\documentclass[10pt,twocolumn,letterpaper]{article}

\usepackage{wacv}
\usepackage{times}
\usepackage{epsfig}
\usepackage{graphicx}
\usepackage{amsmath}
\usepackage{amssymb}
\usepackage{tabularx} 
\usepackage{float}
\newcolumntype{Y}{>{\centering\arraybackslash}X}
\newcolumntype{b}{X}
\newcolumntype{m}{>{\hsize=.37\hsize}X}
\newcolumntype{s}{>{\centering\arraybackslash\hsize=.25\hsize}X}

\wacvfinalcopy 

\def\wacvPaperID{56} 

\ifwacvfinal\fi
\begin{document}

\title{Rotational Rectification Network: \\Enabling Pedestrian Detection for Mobile Vision}
\author{Xinshuo Weng \\
Carnegie Mellon University\\
{\tt\small xinshuow@andrew.cmu.edu}
\and
Shangxuan Wu \\
Carnegie Mellon University\\
{\tt\small shangxuw@andrew.cmu.edu}
\and
Fares Beainy \\
Volvo Construction Equipment\\
{\tt\small fares.beainy@volvo.com}
\and
Kris M. Kitani \\
Carnegie Mellon University\\
{\tt\small kkitani@cs.cmu.edu}
}

\maketitle
\ifwacvfinal\thispagestyle{empty}\fi


\begin{abstract}
Across most pedestrian detection datasets, it is typically assumed that pedestrians will be standing upright with respect to the image coordinate system. This assumption is not always valid for many vision-equipped mobile platforms, such as mobile phones, UAVs, or construction vehicles on rugged terrain. In these situations, the motion of the camera can cause images of pedestrians to be captured at extreme angles. This can lead to inferior pedestrian detection performance when using standard pedestrian detectors. To address this issue, we propose a Rotational Rectification Network (R2N) that can be inserted into any CNN-based pedestrian (or object) detector to adapt it to significant changes in camera rotation. The rotational rectification network uses a 2D rotation estimation module that passes rotational information to a spatial transformer network \cite{Jaderberg2015} to undistort image features. To enable robust rotation estimation, we propose a Global Polar Pooling (GP-Pooling) operator to capture rotational shifts in convolutional features. Through our experiments, we show how our rotational rectification network can be used to improve the performance of state-of-the-art pedestrian detectors under heavy image rotation by up to 45\%.
\end{abstract}

\section{Introduction}
\label{sec:intro}

Pedestrian detection is an active research area in computer vision that rapidly progressed over the past decade. There are many benchmark pedestrian detection datasets available for learning and evaluation \cite{Dollar2012, Geiger2012, eth_biwi_00534, Dalal2005}. One common setting in these datasets is that the camera's y-axis is roughly aligned to the direction of gravity meaning that pedestrians are captured in the vertical direction because pedestrians usually stand upright on the ground. This ``upright assumption'' in benchmark datasets distinguishes pedestrians from many objects in the scene. Much work has been devoted to designing features \cite{Dalal2005, Felzenszwalb2009} or model architectures \cite{Zhang2016, Ouyang_2013_ICCV, Tian2015} to learn the appearance of upright pedestrians.

\begin{figure}[!t]
\includegraphics[width=\linewidth]{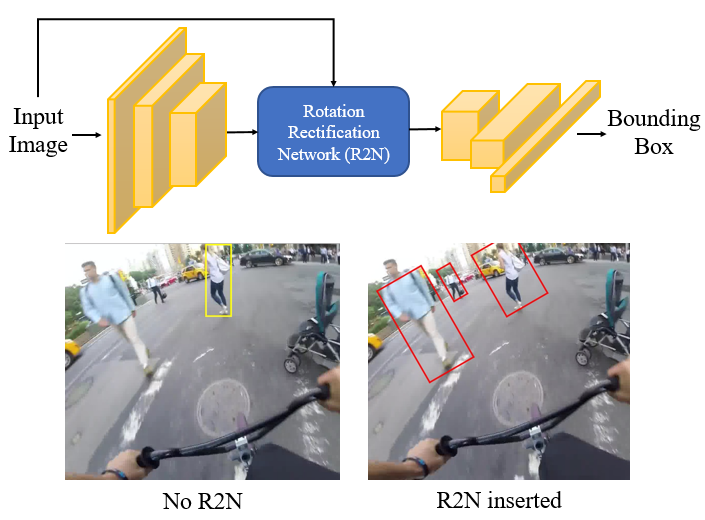}
\caption{ \textbf{Upper:} Schematic diagram of how the proposed rotation rectification network (blue) is inserted into a CNN-based pedestrian detector (yellow). \textbf{Lower:} Illustrative results produced by a state-of-the-art pedestrian detector (\textbf{left}) and with the R2N inserted (\textbf{right}). The R2N increases the robustness of the network to image rotation and decreases the miss-rate of detection.}
\end{figure}
However, this upright assumption may not always be valid in real-world situations where the camera orientation is highly dynamic. For example, when recording a video with a mobile phone camera, the angle of the camera can vary significantly as one walks or runs. For cameras installed on construction vehicles, the upright assumption is easily invalidated when recording video over rough terrain. In both examples, the projection of pedestrians in the image can be at extreme angles of rotation, and detecting pedestrians in such situations is difficult with current state-of-the-art models.

\begin{figure*}[!t]
    \centering
    \includegraphics[width=0.8\linewidth]{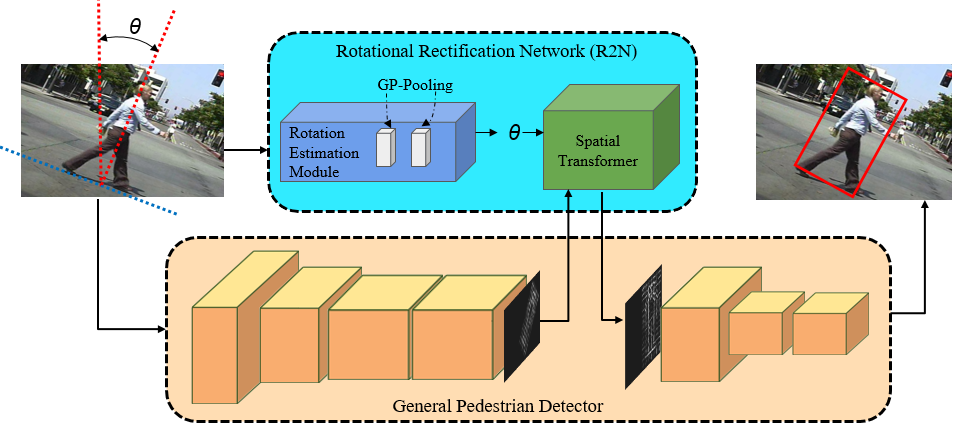}
    \caption{Architectural Overview. Rotation rectification network (R2N) (cyan) is inserted into the intermediate layer of a CNN-based pedestrian detector (yellow). R2N uses a rotation estimation network (see Figure \ref{fig:rotation_estimation}) with GP-Pooling (gray) operators to estimate the rotation angle (blue). The estimated rotation angle $\theta$ is passed to the Spatial Transformer (green). R2N warps the image features to remove global rotation. The last layer (yellow) yields tight rotated bounding boxes.}
    \label{fig:main}
    \vspace{-3mm}
\end{figure*}

One straightforward way to achieve rotational robustness for pedestrian detection is to simply increase the size of the training data to include more instances of pedestrians imaged at an angle. When new data is not available, the existing data may also be augmented \cite{VanDyk2001} by rotating images. While such a data augmentation approach can lead to improvements, merely creating more data does not necessarily address the fundamental problem of understanding and modeling image rotations.

An alternative solution is to infer the rotational distortion of the image and to remove the effect of that distortion prior to detection within a unified rotation-invariant detection network. However, estimating rotational changes in an image is difficult with the current paradigm of convolutional feature extraction because they are based on a rectangular spatial decomposition of the image. In other words, rotational changes in image content can produce very different feature responses in the upper convolutional layers of a convolutional neural network. 

To facilitate a smoother change in convolutional feature responses due to image rotation, we propose the use of a novel Global Polar Pooling (GP-Pooling) operator, which converts rectangular convolutional feature responses into a polar grid system. Using polar coordinates, rotations of the input images result in translational shifts of the polar features maps making it easier for higher-level convolution layers to model image rotation. In this way, our proposed GP-Pooling operator provides CNNs the ability to encode image rotations more effectively. 

To obtain rotational invariance during detection, we propose a rotational rectification network (R2N) that can be flexibly inserted into an intermediate layer of a general detection network. The R2N uses a CNN with GP-Pooling layers to estimate the rotation angle present in the images. Then, the estimated rotation, $\theta$, is passed to a spatial transformer network to undistort the image features. An overview of the network architecture is illustrated in Figure \ref{fig:main}. We show that after removing the effect of rotation inside a network, the general detector can be easily adapted to work on pedestrians imaged at arbitrary rotation angles.

The contributions of our work include the proposals of  (1) a Global Polar Pooling (GP-Pooling) operator, which can be used to encode the radial distribution of features within a general CNN architecture, and (2) a rotational rectification network (R2N) that can be inserted into CNN-based detectors to achieve rotational invariance.

\section{Related Work}
\noindent\textbf{Rotational Robustness in CNNs.} Existing methods to incorporate CNNs with rotational robustness can be split into two categories. The first includes methods to add robustness by manipulating the images or feature maps. Prior works \cite{Cheng2016, Laptev2016, Henriques2016, Gatica-perez, Laptev2015} achieve rotational robustness by augmenting the input images on the fly and fusing the response in the upper layer of the network. Dieleman \etal \cite{Dieleman2016} copy the feature responses from intermediate layers in four $90^{\circ}$ angles and compress them by their proposed operators, which compactly achieves rotational robustness. However, by augmenting the data, these methods are only robust to a discrete set of rotations instead of $360^{\circ}$ continuous rotations. This does not directly address the fundamental problem of incorporating rotation robustness in CNNs.

\cite{Jaderberg2015} introduce a general warp framework called Spatial Transformer Networks to enable affine transformations with differentiable sampling inside the network. It achieves transformation invariance within CNN architectures very efficiently without data augmentation. An important point often overlooked is that the design of the \textit{localization network}, with the purpose to estimate transformation parameters, is not explored in the original work. They use two baseline CNNs as the localization networks in the spatial transformer: (1) two fully-connected layers, and (2) a CNN with two convolutional and two max pooling layers. The models are evaluated on the distorted MNIST dataset, which is small with low-resolution images and does not offer strict criterion to judge transformation invariance of a network. In other words, the design of a network with natural transformation invariance is still an open problem. Our work is complementary to the spatial transformer because our proposed rotation estimation module with GP-Pooling operators can be viewed as an expert localization network with natural rotation invariance.

The second category methods achieve rotational robustness by modifying filters within CNN architectures instead of manipulating the data. Cohen \etal \cite{Cohen2016, Cohen2017} apply kernel-based pooling to sample responses in symmetry space such that only the least important features are lost at each layer. Prior works \cite{Marcos2016, Zhou, Marcos2016texture, And2015} replicate and transform the learned canonical filters in a finite set of orientations and then fuse the output responses at each layer to achieve rotational robustness. Similarly, these methods are robust to only a discrete set of rotations. Instead, our proposed GP-Pooling operator adds rotational robustness to general CNN architectures in $360^{\circ}$ continuous rotations.

Perhaps Harmonic Network (H-Nets) \cite{Worrall} is the closest work to ours. H-Nets replace regular CNN filters with complex circular harmonics and is also able to capture continuous rotational changes. However, H-Nets assume the learned filters are in the harmonic wavelets space whereas GP-Pooling does not impose any assumption on the image filters. Moreover, H-Nets is designed to learn local rotational robust filters while GP-Pooling operator focuses on global rotational changes. More importantly, most existing methods only test rotational robustness on simple tasks, such as digit recognition from MNIST. Our proposed GP-Pooling operator succeeds on real-world tasks, namely, pedestrian detection on the Caltech Pedestrians dataset.

\vspace{2mm}\noindent\textbf{Detection.} Recent detection methods are based on region proposals that perform detection by classifying region proposals of images and regressing the bounding box. For example, Ren \etal \cite{Ren2015} introduce a Region Proposal Network (RPN) to enable nearly cost-free region proposals and propose a unified detection framework. Liu \etal \cite{Liu2015} introduce default boxes, which tiles input images and then regress the offset for each box in the work of Single Shot MultiBox Detector (SSD). In the context of pedestrian detection, Zhang \etal \cite{Zhang2016} analyze the performance of Faster-RCNN on pedestrian detection and propose a simple and powerful baseline for pedestrian detection based on RPN. In many of these methods, the region proposals are represented by axis-aligned rectangles, which are not suitable for detecting pedestrians imaged at an angle. To address this issue, Ma \etal \cite{Ma2017} propose a novel framework to detect text with arbitrary orientation in natural scene images. In their work, they present the Rotation Region Proposal Networks (RRPN) to generate rectangular proposals at different rotations instead of axis-aligned proposals. This approach is limited because the RRPN can only deal with a discrete set of rotations and it is only applicable to proposal-based detection networks.

\begin{figure}[tb]
    \centering
    \includegraphics[width=0.70\linewidth]{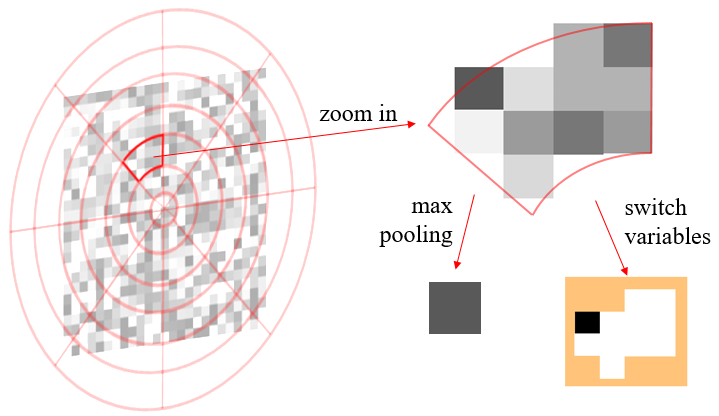}
    \includegraphics[width=0.28\linewidth]{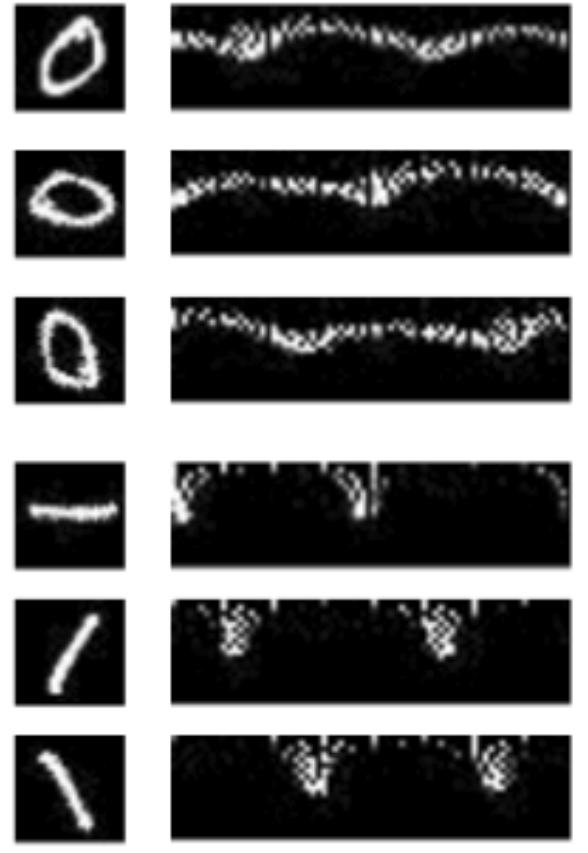}

    \caption{\textbf{Left:} GP-Pooling Operator. Feature map in polar coordinates. \textbf{Right:} Feature Responses of GP-Pooling Operator. Input rotations result in translational shifts of feature responses.}
    \label{fig:RE-Pooling}
    \vspace{-0.3cm}
\end{figure}

\section{Global Polar Pooling (GP-Pooling)}
In a CNN, pooling increases the receptive field and filters out the noisy feature responses from previous layers. Moreover, existing pooling operators, especially spatial max pooling, are frequently used because of their robustness to translation. In other words, translational changes in image content can produce feature responses with translational shifts in existing spatial pooling operators. However, this is not the case for rotational changes. Current paradigms of convolutional feature extraction are strictly based on a rectangular spatial decomposition of the image features. As a result, any rotational changes in image content can produce a very different feature response.

To achieve a smoother change in general CNN feature responses due to rotational changes, we propose the Global Polar Pooling (GP-Pooling) operator. This operator extends existing pooling operators from rectangular to a radial decomposition. It makes the global rotational changes from the input image content easily recognizable in CNNs. Specifically, the GP-Pooling operator represents convolutional feature responses on a polar grid system such that any in-plane rotation from an input image results in a translational shift of the polar feature map. Then, the translational shifts can be easily captured by the upper layers of the network. 

The core idea of how GP-Pooling works is illustrated in Figure \ref{fig:RE-Pooling}. Inside the GP-Pooling operator, the feature maps are represented in a polar coordinate system with the origin defined at the center of the feature map. While our GP-Pooling layer is designed primarily for in-plane rotation about the image center, we empirically find that it can handle moderate levels of off-center rotation. To be concrete, a pixel $P(x, y)$ of the feature map with width $w$ and height $h$ can be represented in polar coordinate $P_p(x_p, y_p)$ by:
\vspace{-0.3cm}
\begin{equation}
    P_n(x_n, y_n) = (x - \frac{w}{2}, -y + \frac{h}{2}) 
\end{equation}
\vspace{-0.6cm}
\begin{equation}
    P_p(x_p, y_p) = (\sqrt{x_n^2 + y_n^2}, \ \textrm{atan2}(y_n, \ x_n)),
\end{equation}
where $P_n(x_n, y_n)$ is the normalized coordinate based on the center of the feature map.

The key difference of GP-Pooling from existing pooling operators is that we define parameters of kernel size, stride, and padding along radial and angular axes in polar coordinates. These parameters determine how the input feature map is tiled into a grid. Inside each grid cell, a max operation is executed for pixels that fall into that cell. A switch variable records the location of maximum activation. Then, the gradient flows back to this location during backpropagation as illustrated in Figure \ref{fig:RE-Pooling}, where the input feature map is tiled with a kernel size of $\frac{\pi}{4}$, stride of $\frac{\pi}{4}$, and padding of $0$ along the angular axis. In this illustrative example, the feature map is tiled into eight angular sectors, each of size $\frac{\pi}{4}$ radians, which are further divided into seven cells along the radial axis. In practice, it is necessary to set the stride and kernel size along the angular axis to $\frac{\pi}{180}$ to capture one degree of image rotation.

To demonstrate the functionality of converting rotational changes to translational shifts, we visualize the output features of the proposed GP-Pooling operator in Figure \ref{fig:RE-Pooling}. We take two MNIST images ($28 \times 28$) as the input of the GP-Pooling operator. The kernel size, stride, and padding along the angular and radial axes are $\frac{\pi}{36}$, $\frac{\pi}{36}$, $0$, $1$, $1$, and $0$, respectively. This results in output features of size $20 \times 72$. We then re-scale these to a size of $28 \times 100$ to obtain the same height as the input image for better visualization. The results show that the feature responses approximately shift leftwards or rightwards when we rotate the input image.

\begin{figure*}[!t]
    \centering
    \includegraphics[width=0.9\linewidth]{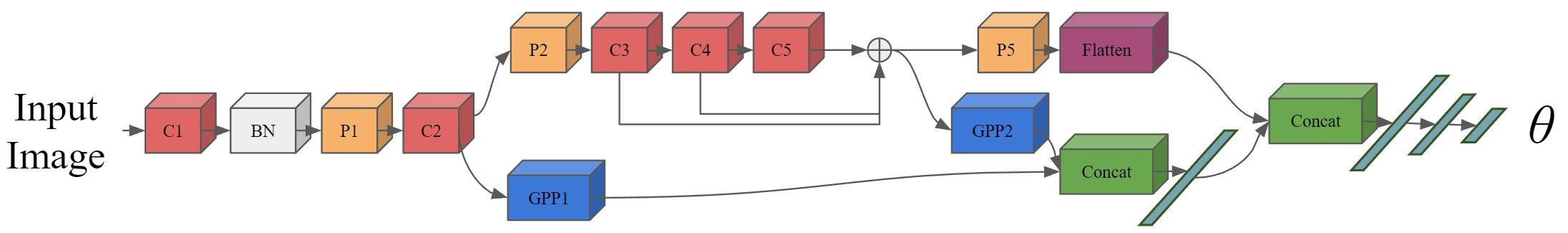}
    \caption{Architecture of a rotation estimation module embedded with the GP-Pooling (blue) operators. This network is composed of convolution (red), max pooling (yellow), GP-Pooling (blue), batch normalization (gray), concatenation (green), flatten (magenta), and fully connected (cyan) layers. The rotation estimation module takes images or image features as the input. The final regression layer produces the estimation of rotation $\theta$ present in the input.}
    \label{fig:rotation_estimation}
    \vspace{-0.3cm}
\end{figure*}


\vspace{2mm}\noindent\textbf{Rotation Estimation Module.} To estimate the rotation parameter from images or image features efficiently, the property of converting rotational changes to translational shifts is beneficial. We insert multiple GP-Pooling layers into the rotation estimation module in a multi-scale manner and concatenate their output feature responses with features from the spatial max pooling layer. The rotation estimation module ultimately outputs the estimation of rotation, $\theta$, ranging from $-\pi$ to $\pi$ present in the input image by solving a regression problem. A typical architecture of our rotation estimation module embedded with the GP-Pooling operators is shown in Figure \ref{fig:rotation_estimation}. 

\section{Rotational Rectification Network (R2N)}

Here we describe the rotational rectification network (R2N) and how we fit it into a general pedestrian detector to achieve rotation-robust detection. The R2N takes two inputs: (1) the input image, and (2) a feature map from an intermediate layer of the detection network. Next, it outputs a warped image feature where the global rotation is removed. This process transforms a complex task of arbitrary-oriented pedestrian detection into an easier task of upright pedestrian detection. The overall architecture is shown in Figure \ref{fig:main}. The R2N is composed of a rotation estimation module and a spatial transformer. We use the estimation of rotation $\theta$ from the rotation estimation module to construct a $2 \times 3$ transformation (rotation) matrix $M$.

\vspace{-0.2cm}
\begin{equation} \label{eq:rot_matrix}
M = \begin{bmatrix}
    \textrm{cos}\theta   & \textrm{-sin}\theta & 0 \\
    \textrm{sin}\theta   & \textrm{cos}\theta & 0 \\
\end{bmatrix}
\end{equation}
\vspace{-0.2cm}

The transformation matrix $M$ is then passed to the spatial transformer to warp the input feature map. This warping removes the effect of global rotation distortion present in the image features prior to detection. In our paradigm, the spatial transformer is used only for image warping. The warp parameters are provided by our specific localization net\footnote{Please refer to \cite{Jaderberg2015} for more details about the spatial transformer and definition of localization net.}, the rotation estimation module, which is designed for increasing robustness to image rotation. 

We emphasize, again, that the R2N is a separate module, independent of the pedestrian detector. It can work as a plugin and be inserted into an intermediate layer in many CNN-based pedestrian detectors to achieve rotation-invariant detection. In practice, we usually insert the R2N module into the feature extraction part of the network (e.g., after the pool3 layer of the VGG part). The R2N can enable the final layer of the pedestrian detector to yield the tight rotated bounding boxes based on the estimated rotation, $\theta$.

\section{Datasets}

To evaluate the performance of our GP-pooling layer and our rotation invariant R2N network, we need a dataset with images of people undergoing heavy rotation. For a detailed quantitative analysis, we utilize rotated MNIST and rotated Caltech datasets, where the digit and pedestrian images are synthetically rotated at various angles. To verify the performance on real rotated data, we obtained a YouTube Wearable Video dataset, where people with wearable cameras capture images during dynamic activities (e.g., running or riding a bike) such that pedestrians are imaged at various angles.

\subsection{Rotated MNIST}

The rotated MNIST dataset is created by rotating images from the MNIST dataset \cite{Liu2003} with the rotation angle uniformly selected from $-90^{\circ}$ to $90^{\circ}$ (the upper half of a circle). The rotated MNIST dataset contains 10,000 training images, 2,000 validation images, and 50,000 testing images, each with a size of 28$ \times $28. We emphasize that this dataset is not used to evaluate digit classification, but to evaluate the performance of image rotation estimation.

\vspace{-0.2cm}
\begin{figure}[!h]
    \centering
    \includegraphics[width=0.10\linewidth]{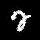}
    \includegraphics[width=0.10\linewidth]{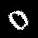}
    \includegraphics[width=0.10\linewidth]{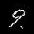}
    \includegraphics[width=0.10\linewidth]{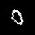}
    \includegraphics[width=0.10\linewidth]{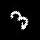}
    \caption{Sample images from the rotated MNIST dataset.}
    \label{fig:rotated_mnist_illustration}
    \vspace{-4mm}
\end{figure}

\subsection{Rotated Caltech} \label{sec:rot_caltech}

We use a rotated version of the Caltech pedestrian dataset to evaluate the ability of a pedestrian detector to detect people imaged at varying angles of rotation. The original Caltech dataset contains six video sequences for training and five for testing. Since consecutive images are very similar, images are sampled every three frames in the training set and every 30 frames in the testing set. This results in 42,786 training images and 4,024 testing images in our rotated Caltech dataset. We rotate all images by uniformly selecting the rotation angle from $-90^{\circ}$ to $90^{\circ}$, which are saved as ground truth for training and testing the rotation estimation module.

\vspace{-0.2cm}
\begin{figure}[!h]
    \centering
    \includegraphics[width=0.20\linewidth]{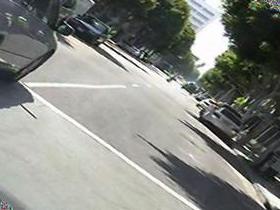}
    \includegraphics[width=0.20\linewidth]{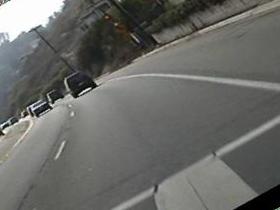}
    \includegraphics[width=0.20\linewidth]{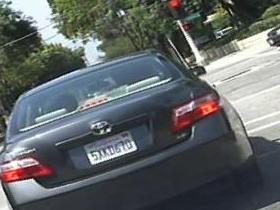}
    \includegraphics[width=0.20\linewidth]{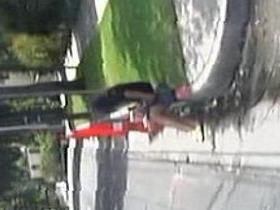}
    \caption{Sample images from the rotated Caltech dataset.}
    \label{fig:rotated_caltech_illustration}
    \vspace{-0.4cm}
\end{figure}

\subsection{YouTube Wearable}
To evaluate the performance of our rotation invariant R2N in the real world, we create the YouTube Wearable dataset where images contain pedestrians with various poses without manual rotation. We obtained 100 short YouTube videos recorded by people with wearable cameras. To have a high possibility to contain pedestrians within images, we cut 4,000 frames from videos where images are taken outdoor in the city. As a result, around 500 images contain pedestrians, and we label the bounding boxes manually and remove boxes under 400 pixels.

\vspace{-0.2cm}
\begin{figure}[!h]
    \centering
    \includegraphics[width=0.30\linewidth, height=0.15\linewidth]{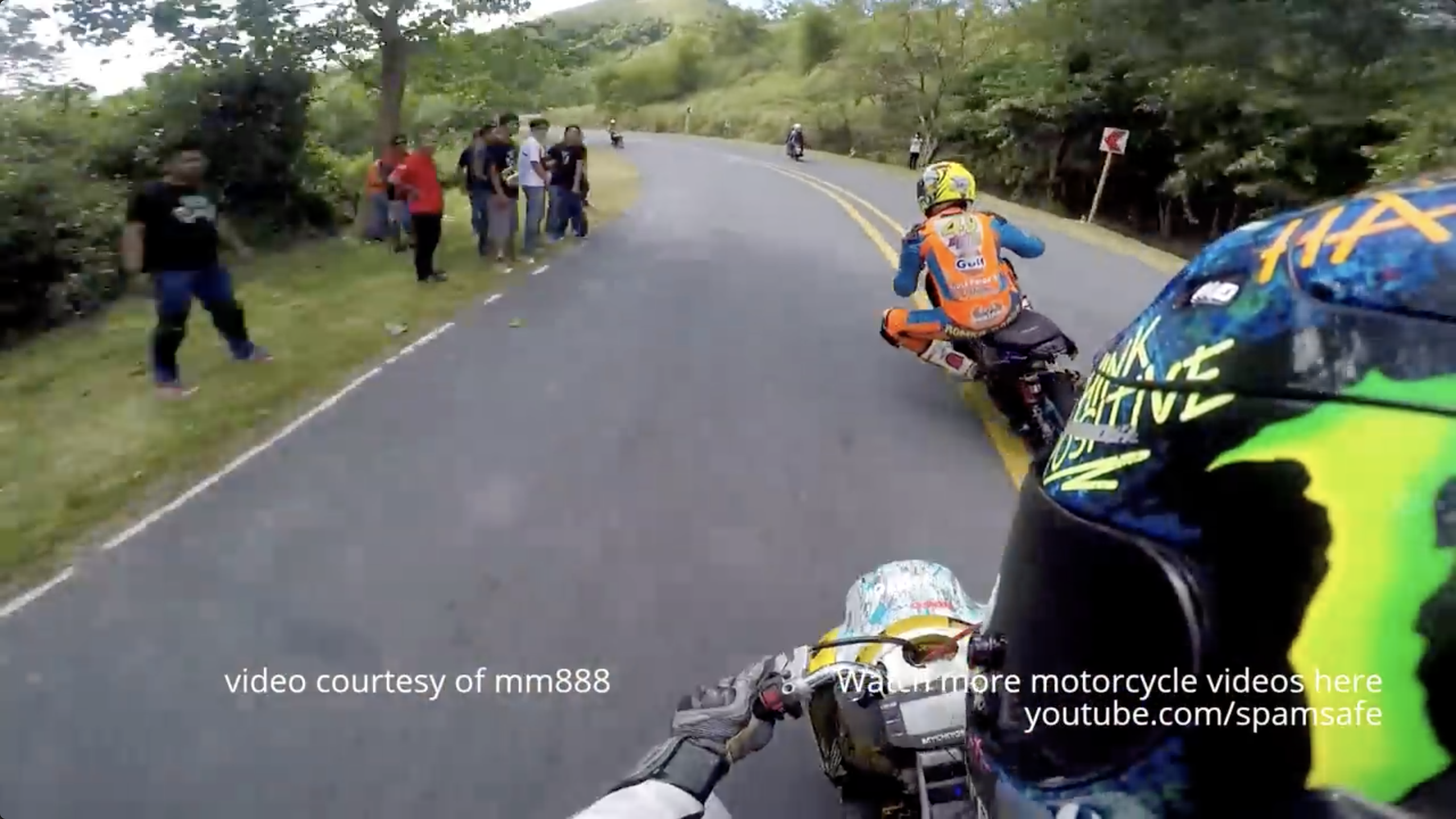}
    \includegraphics[width=0.30\linewidth, height=0.15\linewidth]{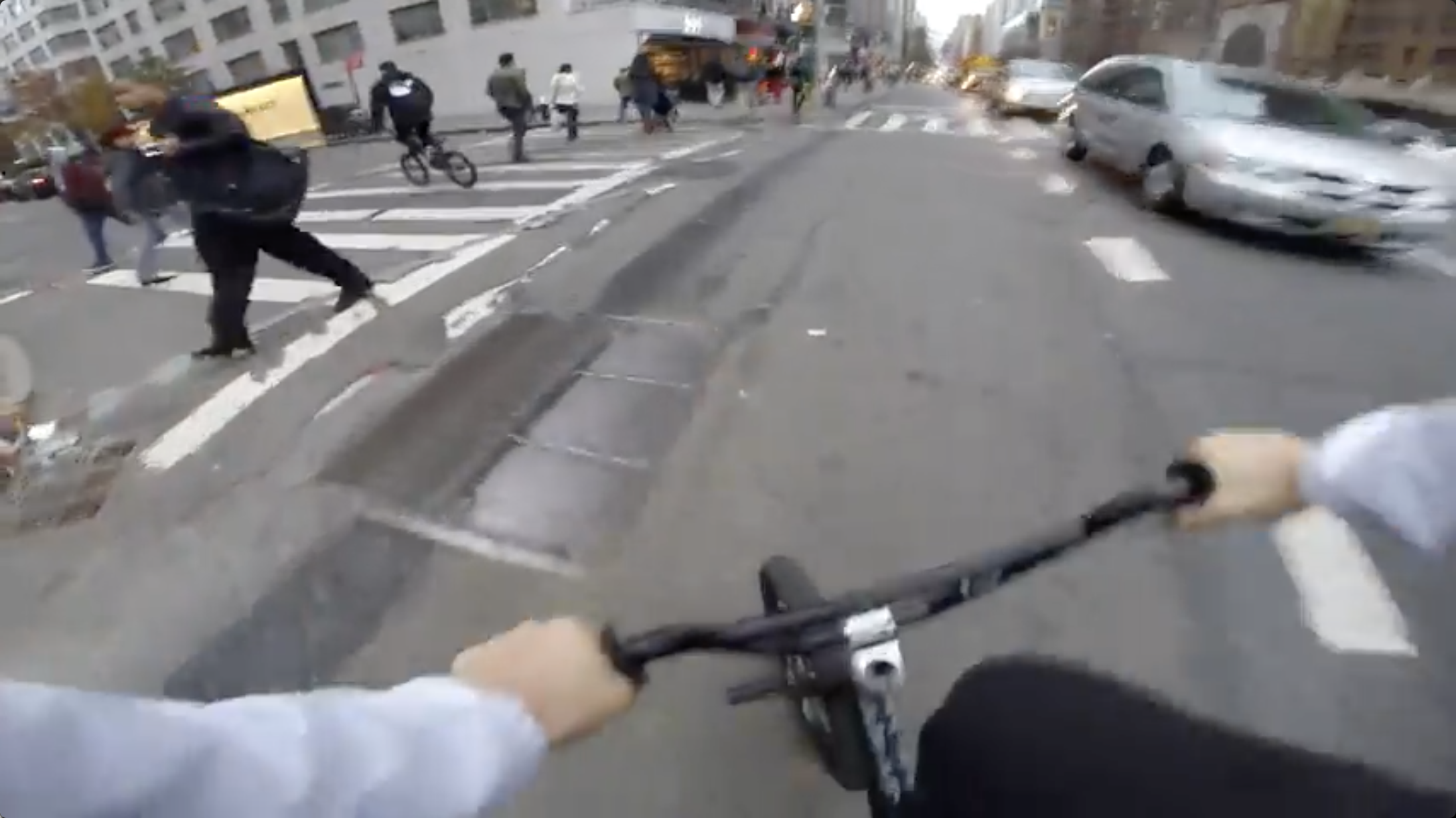}
    \includegraphics[width=0.30\linewidth, height=0.15\linewidth]{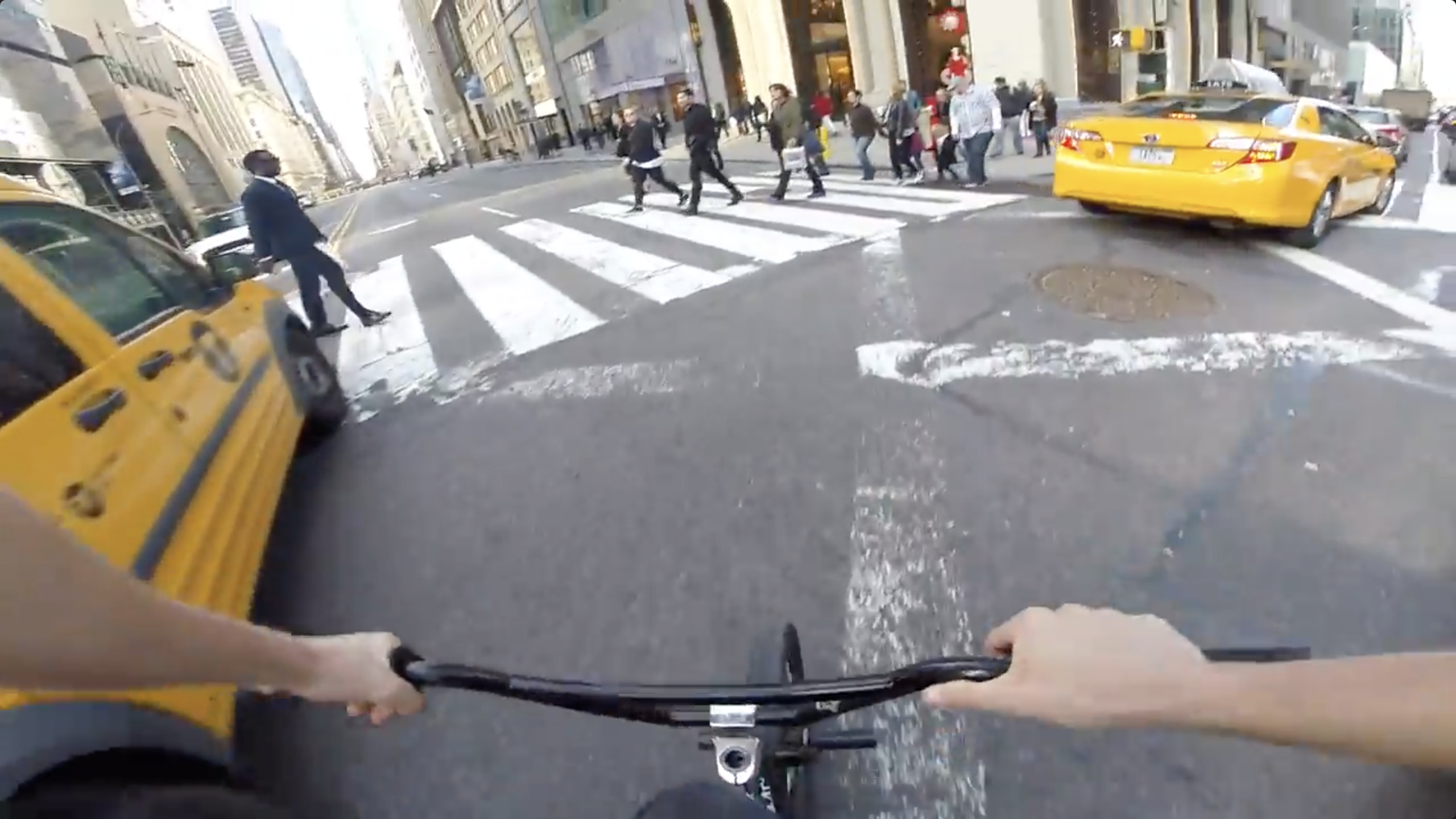}
    \caption{Sample images from the YouTube Wearable dataset.}
    \label{fig:youtube_illustration}
    \vspace{-0.4cm}
\end{figure}

\begin{table*}[tb]
\centering
\footnotesize
\begin{tabularx}{\linewidth}{msb}
        \hline
        \hline
        Name & With STN? \cite{Jaderberg2015} & Description \\
        \hline
        2FC & No & a 2-layer fully connected network (20 and 1 neurons per layer respectively).  \\
        CNN & No & a standard CNN described in Table \ref{tab:topo_mnist}. \\
        \textbf{CNN+GP-Pooling} & No & Two GP-Pooling operators are inserted into the CNN in the 2nd row after the conv1 and conv2 layers before concatenating with the pool2 features. \\
        \hline
        \hline
        STN-2FC \cite{Jaderberg2015} & Yes & a spatial transformer with 2FC as the rotation estimation module.  \\
        STN-CNN \cite{Jaderberg2015} & Yes & a spatial transformer with CNN as the rotation estimation module.  \\
        \textbf{STN-CNN+GP-Pooling} & Yes & a spatial transformer with CNN+GP-Pooling as the rotation estimation module.  \\
        \hline
\end{tabularx}
\vspace{0.05cm}
\caption{Baselines used in two rotated MNIST experiments.}
\label{tab:baselines_rot_mnist}
\vspace{-0.2cm}
\end{table*}

\begin{table*}[tb]
\centering
\footnotesize
\begin{tabularx}{\linewidth}{YYYYYYYYYYYY}
        \hline
        \hline
        Layer & conv1 & ReLU & pool1 & conv2 & ReLU & pool2 & fc3 & ReLU & dropout & fc4 & sigmoid \\
        \hline
        Units & 16 & 16 & 16 & 32 & 32 & 32 & 20 & 20 & 20 & 1 & 1 \\
        Feature & 28$\times$28 & 28$\times$28 & 14$\times$14 & 14$\times$14 & 14$\times$14 & 7$\times$7 & 1 & 1 & 1 & 1 & 1 \\
        \hline
\end{tabularx}
\vspace{0.05cm}
\caption{Topology of the CNN used in the rotated MNIST dataset.}
\label{tab:topo_mnist}
\end{table*}

\section{Evaluating the Rotation Estimation Module}
\label{sec:rem}

The success of the R2N module to enable rotation invariant detection relies heavily on the precision of the estimated rotation, $\theta$, so it is critical to estimate the image rotation precisely. In the first experiment, we evaluate the accuracy of the rotation estimation module using the proposed GP-Pooling operator on rotated MNIST and rotated Caltech dataset.

\subsection{Rotation Estimation on Rotated MNIST} \label{sec:rot_mnist}
To see how the rotation estimation module works independently, despite the R2N, we evaluate three baseline rotation estimation modules. Descriptions of the baselines are in Table \ref{tab:baselines_rot_mnist} (\textbf{first three rows}). We train all baselines using the rotation angle as the ground truth and evaluate them based on root of sum of squared error.

\vspace{1mm}\noindent\textbf{Training Details.} 
We use Euclidean loss during training. The training takes 160 epochs with a batch size of 128 for all models. Adam optimizer is used with a learning rate of 0.001 and two momentums of 0.9 and 0.999. 

\vspace{1mm}\noindent\textbf{Results.} In Table \ref{tab:exp2_mnist}, we see that the 2-layer fully connected network cannot work very well because the network has too few neurons and a simple structure. Importantly, the GP-Pooling+CNN outperforms the CNN by achieving 22\% lower error (from $11.26^{\circ}$ to $8.78^{\circ}$), showing that adding GP-Pooling operators to the network increases robustness to image rotation. 

\begin{figure}[tb]
    \centering
    \includegraphics[width=0.9\linewidth]{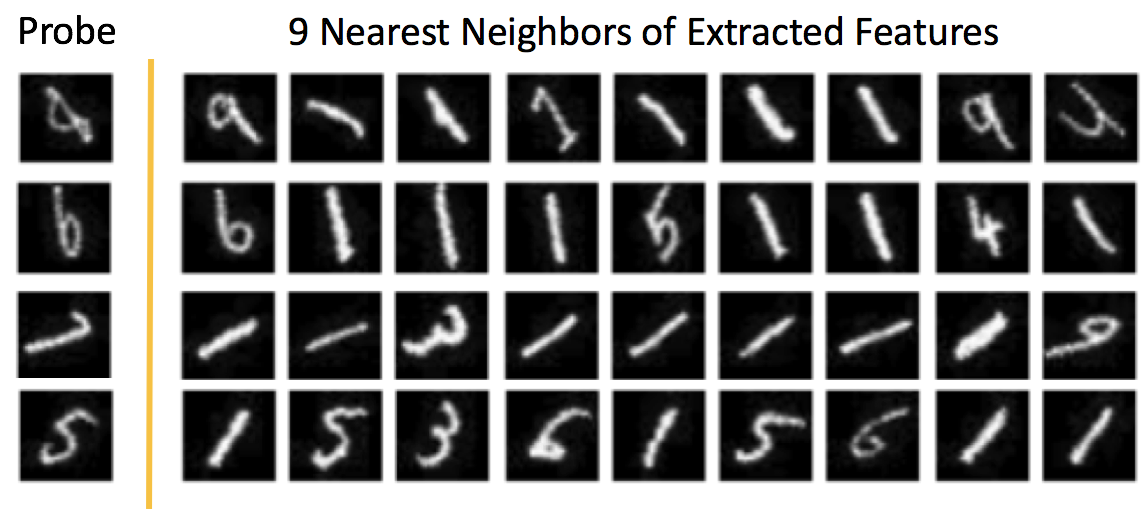}
    \caption{Feature Space Visualization. Nearest neighbors of probe images (left column). Feature representation encodes digit angle, not digit label.}
    \label{fig:nn_vis_mnist}
    \vspace{-0.3cm}
\end{figure}

\vspace{1mm}\noindent\textbf{Visualization.} To understand what is learned in the rotation estimation module (GP-Pooling+CNN), we extract the feature responses before the fc3 layer as a vector representation for each image patch. For a set of seven random digit images, we find the 19 nearest neighbors (Figure \ref{fig:nn_vis_mnist}) and observe that the nearest neighbors are \textit{not} necessarily the same digit. More importantly, the nearest neighbors have the same rotation angle as the query image, which seems to indicate that the learned feature representation encodes digit angle instead of digit label.

\subsection{Rotation Estimation with Spatial Transformer Network (STN) on Rotated MNIST} \label{sec:stn_rot_mnist}
As our proposed rotation estimation module with the GP-Pooling operator is similar and complementary to the spatial transformer (STN), we evaluate the STN with and without our GP-Pooling to see if it is helpful in STN. We use the same rotation estimation modules evaluated in section \ref{sec:rot_mnist}. Descriptions of the baselines are in Table \ref{tab:baselines_rot_mnist} (\textbf{last three rows}). As we estimate the rotation parameters, the transformation matrix of the STN is restricted to a rotation matrix in the form of Equation \ref{eq:rot_matrix}. We train three baseline STNs using digit labels as the ground truth because STN is trained with a digit classification task in the MNIST dataset. However, during testing, to see how the GP-Pooling operator affects the performance of rotation estimation, we do not care about the predicted digit and, instead, evaluate the estimation of rotation produced by the rotation estimation module, an intermediate output of the overall network. 

\vspace{1mm}\noindent\textbf{Training Details.} 
As we are solving a classification task during training, we use Cross Entropy Loss instead of Euclidean Loss. All other training details are the same as in section \ref{sec:rot_mnist}. 
 
\vspace{1mm}\noindent\textbf{Results.} In table \ref{tab:exp1_mnist}, we observe, while STNs are trained with a classification task without using rotation angle, the rotation estimation module is learning to estimate the image rotation. A stronger rotation estimation module (STN-CNN) can have lower estimation error than a simpler model (STN-2FC). Additionally, we found that adding the GP-Pooling operators to the rotation estimation module helps improve the accuracy of rotation estimation even with the spatial transformer.

\begin{table}[tb]
\footnotesize
\centering
\begin{tabular}{ |c|c| }
        \hline
        Methods & Error (degree) \\
        \hline
        \hline
        2FC           &       $44.91^{\circ}$               \\
        CNN           &   $11.26^{\circ}$         \\
        \textbf{GP-Pooling+CNN}   &   $\mathbf{8.78^{\circ}}$    \\
        \hline
\end{tabular}

\vspace{0.2cm}
\caption{Rotation estimation error on rotated MNIST dataset. 2FC, CNN, and GP-Pooling+CNN are the rotation estimation modules defined in Table \ref{tab:baselines_rot_mnist}.}
\label{tab:exp2_mnist} 
\end{table}

\begin{table}[tb]
\centering
\footnotesize
\begin{tabular}{ |c|c| }
        \hline
        Methods & Error (degree) \\
        \hline
        \hline
        STN-2FC \cite{Jaderberg2015}       &   $23.37^{\circ}$        \\
        STN-CNN \cite{Jaderberg2015}       &   $18.00^{\circ}$        \\
        \textbf{STN-GP-Pooling+CNN}                 &   $\mathbf{16.38^{\circ}}$       \\
        \hline
\end{tabular}
\vspace{0.2cm}
\caption{Rotation estimation error on the Rotated MNIST dataset with the Spatial Transformer Network.}
\label{tab:exp1_mnist} 
\end{table}

\subsection{Rotation Estimation on Rotated Caltech}
As the rotated MNIST dataset is very simple and contains only low-resolution images, it cannot be a strict dataset to judge the rotation invariance of a network. As such, we also evaluate the rotation estimation module on the rotated Caltech dataset, where images contain much higher resolution (480 $\times$ 640) and more complex content compared to the MNIST. However, the task of pedestrian detection on the Caltech dataset is beyond the scope of the spatial transformer, so the comparison experiments with STN are not applicable. On the other hand, the task of horizontal line detection is very similar to the rotation estimation (estimation of rotation is the slope of the horizontal line), so we compare our rotation estimation module with two state-of-the-art horizontal line detection algorithms. 

\vspace{1mm}\noindent\textbf{Baselines.}
\vspace{-0.2cm}
\begin{enumerate}
    \setlength\topsep{-1mm}
    \setlength\itemsep{-1.5mm}
    \item \textbf{Zhai \etal \cite{Zhai2016}}: a state-of-the-art CNN-based horizontal line detector.
    \item \textbf{Lezama \etal \cite{Lezama2014}}: an edge-based horizontal line detector.
    \item \textbf{VGG-S \cite{Simonyan2014}}: the small version of VGG network.
    \item \textbf{GP-Pooling+VGG-S}: two GP-Pooling operators are inserted into VGG-S after the conv2 and conv5 layers before concatenation. The topology of the network is shown in Figure \ref{fig:rotation_estimation}. All convolution layers have kernel size of 3, stride of 1, and padding of 1. All max pooling layers have kernel size of 2 and stride of 2.
    \end{enumerate}
\vspace{-0.2cm}

\begin{table}[tb]
\centering
\footnotesize
\begin{tabular}{ |c|c| }
        \hline
        Methods & Error (degree) \\
        \hline
        \hline
        Zhai et al \cite{Zhai2016}          &       $38.79^{\circ}$       \\
        Lezama et al \cite{Lezama2014}      &       $29.26^{\circ}$       \\
        VGG-S \cite{Simonyan2014}           &       $18.33^{\circ}$       \\
        \textbf{GP-Pooling+VGG-S}                    &       $\mathbf{15.82^{\circ}}$       \\
        \hline
\end{tabular}
\vspace{0.2cm}
\caption{Results from the rotated Caltech dataset. The topology of the GP-Pooling+VGG-S is described in Figure \ref{fig:rotation_estimation} and Section \ref{sec:rot_caltech}.}
\label{tab:res_caltech} 
\vspace{-0.3cm}
\end{table}

\noindent\textbf{Training Details.} For baselines 1 and 2, we follow the training procedure from the original work. For baselines 3 and 4, We fine-tune both networks from the VGG-S model pre-trained on ImageNet (up to the pool5 layer) for 16 epochs with a batch size of 4, learning rate of 0.000001, and weight decay of 0.00005.

\vspace{1mm}\noindent\textbf{Results.} In Table \ref{tab:res_caltech}, we observe that \cite{Zhang2016} and \cite{Lezama2014} are not working as well as VGG-S and GP-Pooling+VGG-S because both two horizontal line detection algorithms heavily rely on the geometric priors, which are not always true on the rotated Caltech dataset, especially when the image rotation is heavy. Importantly, when comparing GP-Pooling+VGG-S with VGG-S, we can achieve lower error in degree by 13.7\% (from $18.3^{\circ}$ to $15.8^{\circ}$) by simply adding two GP-Pooling operators to the rotation estimation module. This demonstrates again that the proposed GP-Pooling operator increases robustness to image rotation and improves rotation estimation.

\section{Evaluating the Rotational Rectification Network (R2N)}
\label{sec:r2n}
To see how our R2N can enable rotation invariance in CNN-based pedestrian detectors, we evaluate the performance of two end-to-end pedestrian detectors with R2N on the original and rotated Caltech and YouTube Wearable datasets.

\vspace{1mm}
\noindent\textbf{Baselines.} Faster-RCNN\footnote{The Faster-RCNN used is pre-trained on the VOC2007 dataset. We only evaluate the class of ``person'' from the total 20 classes in VOC2007.} \cite{Ren2015} and RPN-BF\footnote{For RPN-BF, we do not use the random forest part compared to the original work.} \cite{Zhang2016} are very strong pedestrian detectors, and we use four variants as the baselines for comparative analysis: 

\vspace{-0.3cm}
\begin{enumerate}
    \setlength\topsep{-1.5mm}
    \setlength\itemsep{-1.5mm}
    \item \textbf{Base\footnote{Base represents one of the pedestrian detectors: Faster-RCNN or RPN-BF.}}: Faster-RCNN is trained on the VOC2007 dataset, and RPN-BF is trained on the original Caltech dataset.
    \item \textbf{Base+Aug}: Faster-RCNN and RPN-BF are fine-tuned on the mixture of original and rotated Caltech datasets for data augmentation.
    \item \textbf{Base+R2N}: The Faster-RCNN is trained on the VOC2007 dataset and RPN-BF is trained on the original Caltech dataset, then the proposed R2N module is inserted after the pool3 layer of both detectors without fine-tuning on the rotated Caltech dataset.
    \item \textbf{Base+R2N+GT}: Instead of estimating the rotation from the rotation estimation module, the ground truth of rotation is passed to the STN such that all pedestrians within images lead to an upright pose, which should achieve the upper-bound performance of the Base+R2N.
    \end{enumerate}
\vspace{-0.2cm}

The evaluation metric for all following experiments is the log-average miss rate on false positive per image (FPPI) \cite{Dollar2012}. As is standard practice, an intersection over union (IoU) of 0.5 is used to determine true positives. 
\begin{figure*}[!t]
\centering
    \includegraphics[width=0.3\linewidth, height=0.2\textheight]{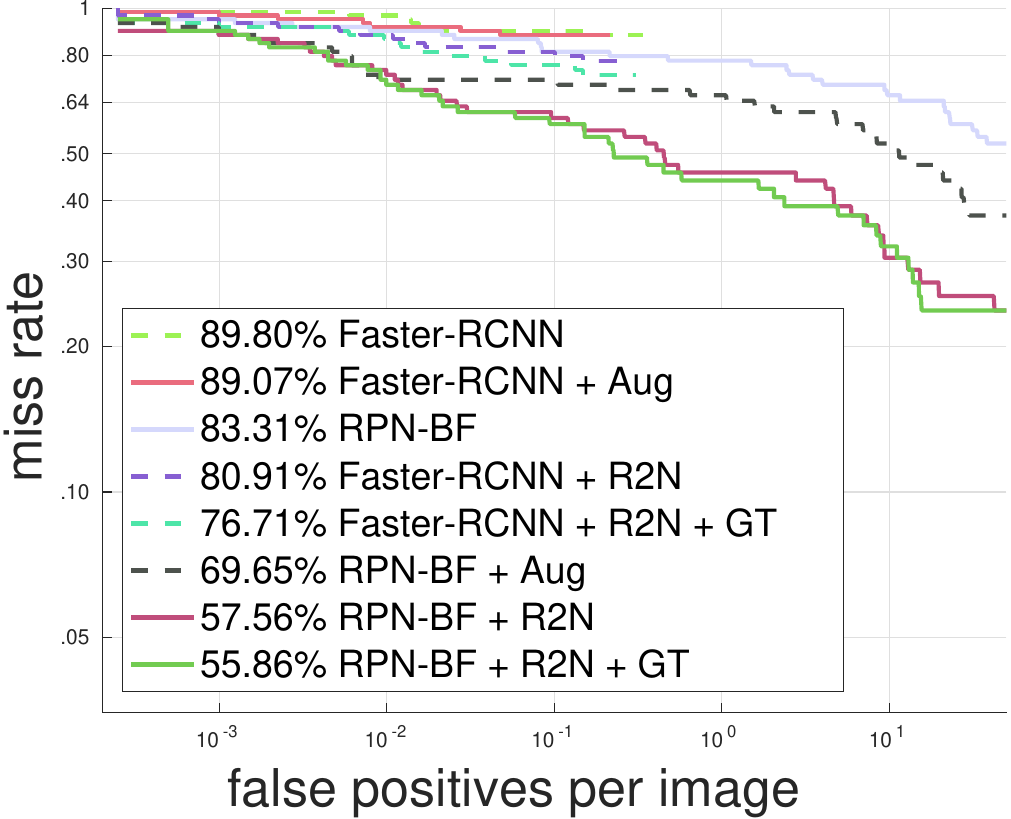}
    \includegraphics[width=0.3\linewidth, height=0.2\textheight]{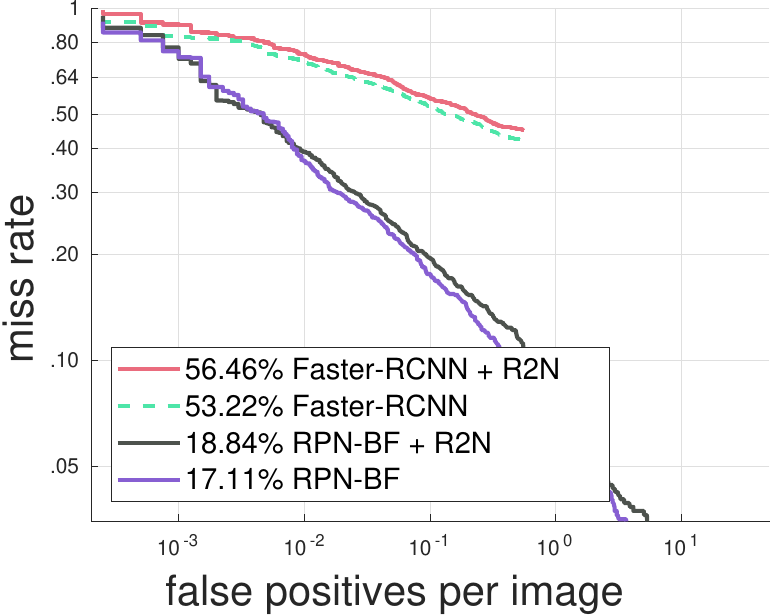}
    \includegraphics[width=0.3\linewidth, height=0.2\textheight]{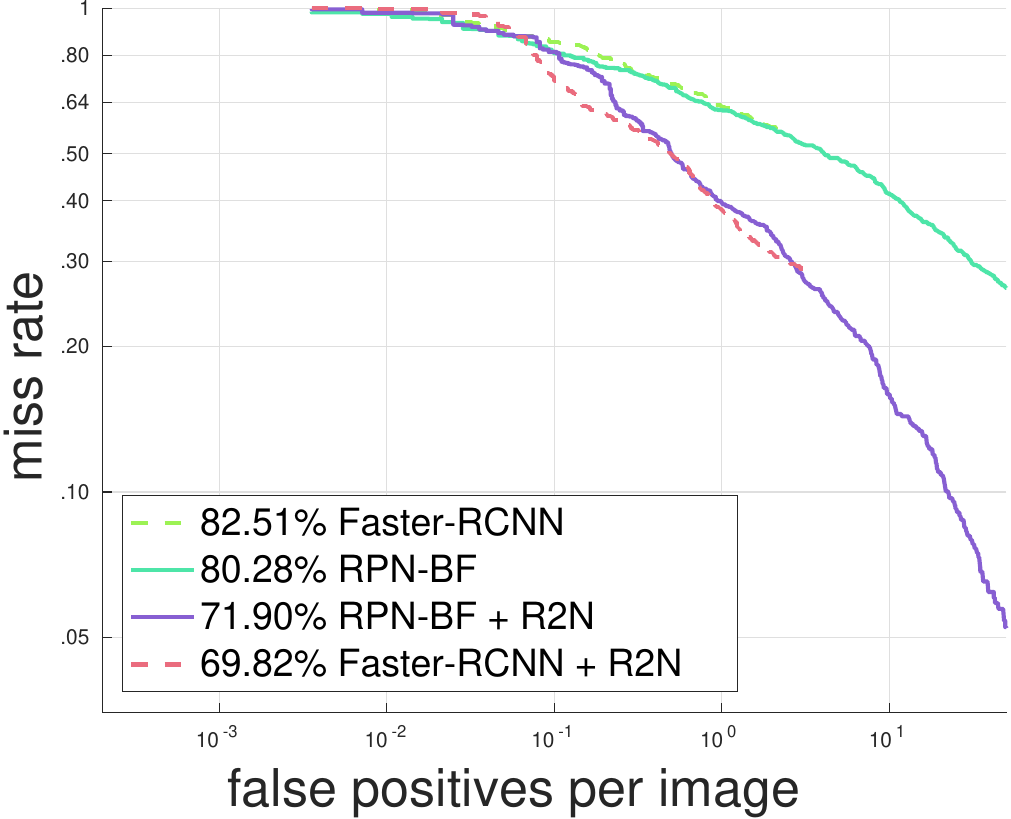}
    \caption{Quantitative results on the rotated Caltech (\textbf{left}), original Caltech (\textbf{middle}), and YouTube Wearable (\textbf{right}) datasets. An Intersection of Union (IoU) of 0.5 is used to determine true positives.}
    \label{fig:quan}
\end{figure*}
\vspace{0.2cm}

\begin{figure*}[!t]
    \centering
    \includegraphics[width=0.99\linewidth]{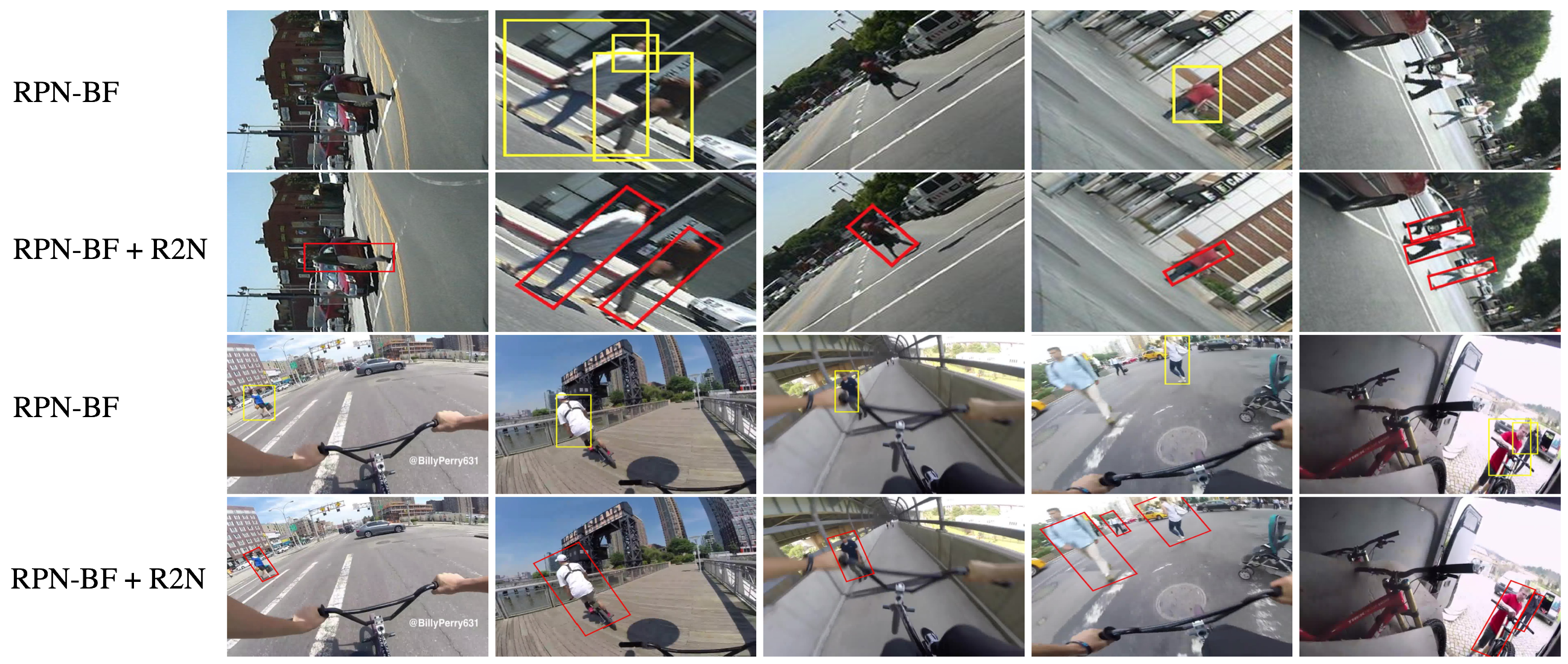}
    \caption{Detection Results. Upper two rows: rotated Caltech dataset. Lower two rows: YouTube Wearable dataset.}
    \label{fig:qua}
    \vspace{-0.2cm}
\end{figure*}

\vspace{1mm}
\noindent\textbf{Results on Rotated Caltech.} We now evaluate the ability of our proposed R2N to transform a pre-trained pedestrian detector into a rotation-invariant pedestrian detector. We begin with experiments on the rotated Caltech dataset with results shown in Figure \ref{fig:quan} \textbf{(left)}. The average miss rate (FPPI) of the base detectors (Faster-RCNN and RPN-BF) have a maximum of 89.8\% and 83.3\%, respectively, because both detectors are only trained on datasets without rotated pedestrians. By fine-tuning the detectors on a mixture of the original and rotated Caltech datasets, the performance of Faster-RCNN and RPN-BF increases 1\% and 20\%, respectively, which is expected because data augmentation is known to improve performance. 

If we add the R2N to each detector, the performance increases by a significant percentage: 11.0\% for Faster-RCNN and 44.6\% for RPN-BF (the average miss rate decreased from 83.3\% to 57.6\%). This result demonstrates that adding rotation invariance via our R2N module to CNNs is more effective at improving detection performance than data augmentation. We emphasize that we do not fine-tune the two baseline models after adding the R2N module. This shows that the R2N module works like a plugin without additional tuning. 

Additionally, when comparing baseline base+R2N with base+R2N+GT, the performance is very close meaning that, in the case of no fine-tuning, the performance of base+R2N nearly achieves the upper bound on the rotated Caltech dataset. Qualitative results are shown in Figure \ref{fig:qua}.

\vspace{2mm}\noindent\textbf{Results on Original Caltech.} To see how adding the R2N module affects the performance of the original pedestrian detector when detecting upright pedestrians, the second experiment focuses on the original dataset instead of the rotated one. When evaluated on the original Caltech dataset (Figure \ref{fig:quan} \textbf{middle}), we found that the performance drops 1\% and 3\% for RPN-BF and Faster-RCNN, respectively, after inserting the R2N into the original detector. This drop is reasonable because we do not jointly fine-tune the networks after inserting the R2N module and the estimation of rotation from the rotation estimation module is not perfectly precise (i.e., it might add some rotations to the upright pedestrians). The performance of the Faster-RCNN variants is much worse than the RPN-BF variants because Faster-RCNN is pre-trained on the VOC2007 dataset and not Caltech dataset.

\noindent\textbf{Results on YouTube Wearable.} To see how our R2N performs on real-world rotated image data, we compare the Faster-RCNN and RPN-BF with and without our R2N on the YouTube Wearable dataset. Quantitative and qualitative results are shown in Figure \ref{fig:quan} (\textbf{right}) and \ref{fig:qua}, respectively. Compared to the results from the Caltech dataset, Faster-RCNN+R2N performs much better, which might result from a similar appearance and scale of pedestrians between the YouTube Wearable and VOC2007 datasets. 
More importantly, adding the R2N increases the performance by 18.2\% for Faster-RCNN and 11.7\% for RPN-BF, although we do not fine-tune the detectors on this completely new dataset. 
This demonstrates again that the proposed R2N can add rotation invariance to a detection network immediately without joint fine-tuning such that the detector can detect pedestrians with various poses. 

\vspace{-1mm}
\section{Conclusion}
\vspace{-1mm}
We introduce the GP-Pooling operator to convert rotational changes to translational shifts enabling CNNs to encode rotational information. We propose a rotational rectification network (R2N) and apply it to a real application of oriented pedestrian detection. We show that the use of R2N can immediately help achieve rotation invariance without any fine-tuning given a detector trained on datasets with only upright pedestrians. 

\vspace{3mm}\noindent\textbf{Acknowledgement.} This work was sponsored in part by JST CREST grant (JPMJCR14E1), NSF NRI grant (1637927) and Volvo Construction Equipment.

{\small
\bibliographystyle{ieee}
\bibliography{egbib}
}

\end{document}